\documentclass[letterpaper]{article} 
\usepackage{aaai2026}  
\usepackage{times}  
\usepackage{helvet}  
\usepackage{courier}  
\usepackage[hyphens]{url}  
\usepackage{graphicx} 
\usepackage{natbib}  
\usepackage{caption} 
\usepackage{algorithm}
\usepackage{algorithmic}
\usepackage{amsmath}
\usepackage{amsfonts}
\usepackage{booktabs}
\usepackage{tabularx}
\usepackage{xcolor}
\usepackage[table]{xcolor}
\usepackage{newfloat}
\usepackage{listings}
\DeclareCaptionStyle{ruled}{labelfont=normalfont,labelsep=colon,strut=off} 
\floatstyle{ruled}
\newfloat{listing}{tb}{lst}{}
\floatname{listing}{Listing}
\title{Replacing Parameters with Preferences: \\Federated Alignment of Heterogeneous Vision-Language Models}
\author {
    Shule Lu\textsuperscript{\rm 1,\rm 2},
    Yujing Wang\textsuperscript{\rm 1,\rm 2},
    Hainan Zhang\textsuperscript{\rm 1,\rm 2}
    Xiaoshan Yang\textsuperscript{\rm 3,\rm 4},
    Hongwei Zheng\textsuperscript{\rm 5},
    Yongxin Tong\textsuperscript{\rm 2},
    Changsheng Xu\textsuperscript{\rm 3,\rm 4},
    Zhiming Zheng\textsuperscript{\rm 1,\rm 2},
}
\affiliations {
    \textsuperscript{\rm 1}Beijing Advanced Innovation Center for Future Blockchain and Privacy Computing\\
    \textsuperscript{\rm 2}Institute of Artificial Intelligence, Beihang University, China\\
    \textsuperscript{\rm 3}MAIS, Institute of Automation, Chinese Academy of Sciences, China\\
    \textsuperscript{\rm 4}School of Artificial Intelligence, University of Chinese Academy of Sciences, China\\
    \textsuperscript{\rm 5}Institute of Artificial Intelligence and Blockchain, Guangzhou University\\
}

\usepackage{bibentry}

\begin{document}

\maketitle

\begin{abstract}
Vision-Language Models (VLMs) have broad potential in privacy-sensitive domains such as healthcare and finance, yet strict data-sharing constraints render centralized training infeasible. Federated Learning mitigates this issue by enabling decentralized training, but practical deployments face challenges due to client heterogeneity in computational resources, application requirements, and model architectures. 
Under extreme model and data heterogeneity, replacing parameter aggregation with preference-based collaboration offers a more suitable interface, as it eliminates the need for direct parameter or data exchange.
Motivated by this, we propose MoR, a federated alignment framework that combines GRPO with Mixture-of-Rewards for heterogeneous VLMs. In MoR, each client locally trains a reward model from local preference annotations, capturing specific evaluation signals without exposing raw data.
To combine these heterogeneous supervision signals, MoR introduces a Mixture-of-Rewards mechanism with learned routing, which adaptively fuses client reward models according to the input and alignment objective. The server then optimizes a base VLM using GRPO with a KL penalty to a reference model, enabling preference alignment without requiring client models to share architectures or parameters. Experiments on diverse public vision-language benchmarks demonstrate that MoR consistently outperforms federated alignment baselines in generalization and cross-client adaptability. Our approach provides a scalable solution for privacy-preserving alignment of heterogeneous VLMs under federated settings.
\end{abstract}


\section{Introduction}
\begin{figure}[!htbp]
    \centering
    \includegraphics[width=0.483\textwidth]{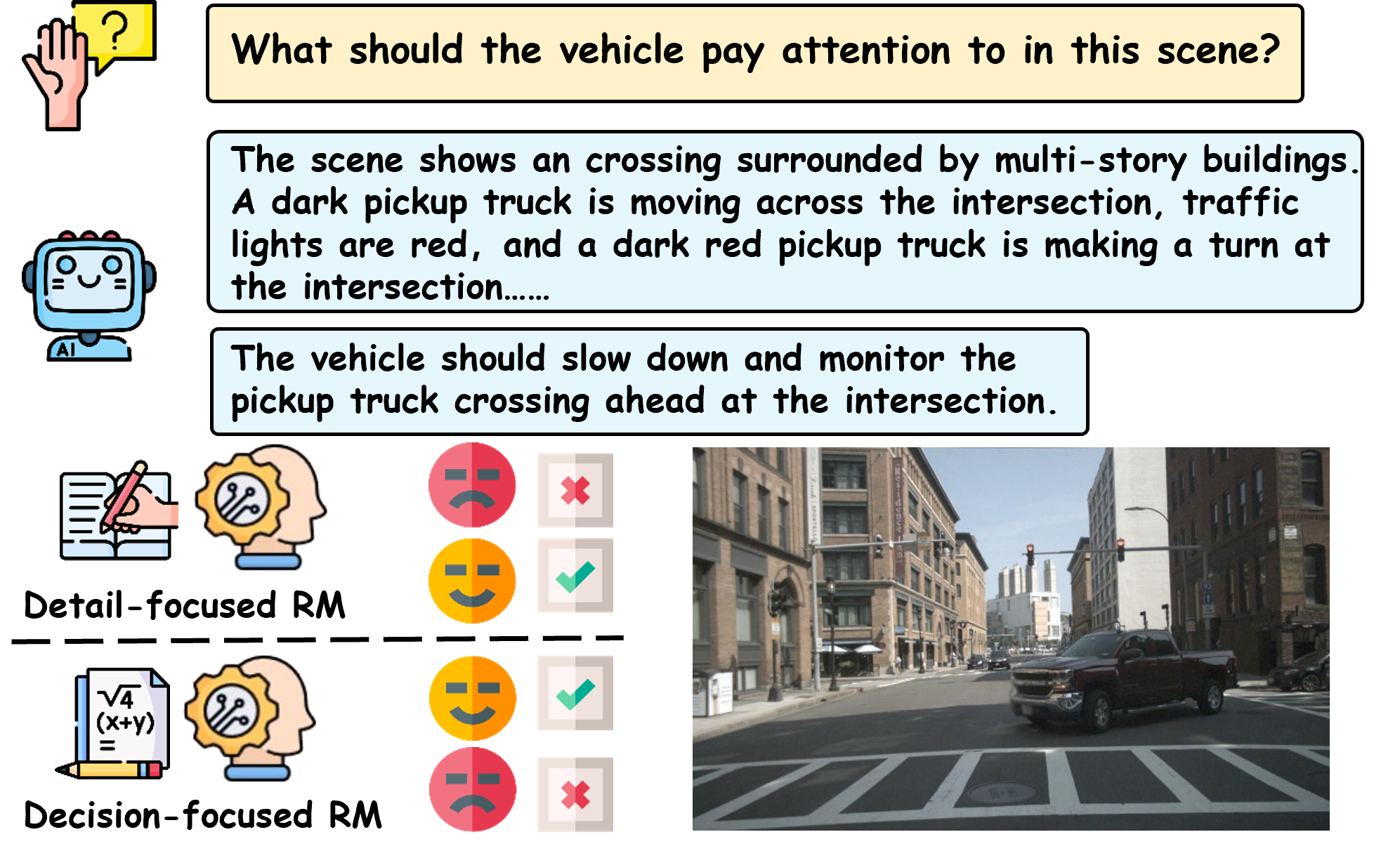} 
    \caption{Illustration of preference heterogeneity in federated VLM alignment. For the same driving image and query, a detail-focused client prefers a comprehensive description, while a safety-focused client prefers a concise hazard-aware response, leading to conflicting reward signals for a single global reward model.}
    \label{fig:data_distribution}
\end{figure}

Vision--language models (VLMs) have demonstrated remarkable capabilities across a wide range of applications \cite{tian2024drivevlm,bordes2024introduction}, including medical image analysis \cite{hartsock2024vision} and decision support in finance \cite{nie2024survey}, where alignment with domain-specific norms, expert preferences, and safety requirements is particularly critical. However, standard alignment methods such as Group Relative Policy Optimization (GRPO) \cite{shao2024deepseekmath} typically assume centralized access to preference supervision for reward modeling or policy optimization. In privacy-sensitive real-world deployments, such supervision often encodes proprietary evaluation criteria, expert judgment, and institution-specific decision boundaries, making direct centralization undesirable or even infeasible. This creates a fundamental tension between the need for high-quality alignment and the distributed, privacy-constrained nature of many practical VLM applications.

Federated learning (FL) \cite{konevcny2016federated,mcmahan2017communication} offers a natural paradigm for privacy-sensitive VLM deployment by keeping raw data local. Nevertheless, federated VLM alignment is substantially more challenging than standard federated optimization because client heterogeneity arises at multiple levels, including data distributions, reward model architectures, and downstream tasks. In alignment settings, such heterogeneity directly translates into heterogeneous reward criteria: different clients may prefer different responses to the same multimodal input depending on their domain knowledge, safety requirements, or task-specific objectives. Consequently, a central challenge is how to preserve and coordinate diverse client-specific reward signals during collaborative training. Existing methods typically rely on static global aggregation, such as parameter averaging \cite{mcmahan2017communication} or unified reward modeling \cite{zhao2023group,ramesh2024group,chakraborty2024maxmin}, which can obscure client specialization and struggle to adapt to input-dependent preference variation. In addition to this heterogeneity issue, real-world federated deployment also requires careful consideration of communication efficiency and privacy robustness.

To address this issue, we seek a reward coordination mechanism that can preserve client-specific specialization while enabling effective collaboration under heterogeneous federated settings. Prior decentralized alignment methods typically rely on shared predictor structures or homogeneous reward assumptions \cite{srewa2025pluralllm,srewa2025systematic}, which limit their ability to handle diverse client objectives and structurally heterogeneous reward models. Inspired by the success of routing-based and mixture-of-experts architectures \citep{jacobs1991adaptive,shazeer2017outrageously,lepikhin2020gshard} in handling heterogeneous experts, we propose \textbf{MoR}, a federated reinforcement alignment framework for VLMs based on \emph{Mixture-of-Rewards}. In MoR, each client maintains a local reward model that captures its own preference criteria, while a lightweight routing mechanism adaptively selects or combines reward signals across clients to produce a context-aware mixed reward. The resulting routed reward is then used to guide GRPO-based policy optimization, with online router updating to track distribution shifts as the policy evolves during training. This design enables MoR to better accommodate heterogeneous clients than static aggregation-based federated alignment methods.

Experimental results on three diverse public vision-language benchmarks show that MoR consistently outperforms existing reward modeling methods and federated learning baselines in terms of generalization, robustness, and cross-client adaptability. Additional privacy attack evaluation and communication-efficiency analysis further demonstrate its practicality for privacy-preserving federated alignment. 
The main contributions of this paper are as follows:
\begin{itemize}
    \item We identify heterogeneous reward coordination as a central challenge in federated VLM alignment, where client heterogeneity arises across data distributions, reward model architectures, and downstream tasks.
    \item We propose MoR, a federated reinforcement alignment framework that uses routing-based Mixture-of-Rewards to adaptively coordinate client-specific reward signals for GRPO-based policy optimization.
    \item We conduct extensive experiments under heterogeneous client settings and show that MoR consistently improves generalization, robustness, and cross-client adaptability, while also demonstrating favorable privacy and communication properties.
\end{itemize}

\begin{figure*}[!ht] 
    \centering
    \includegraphics[width=0.75\textwidth]{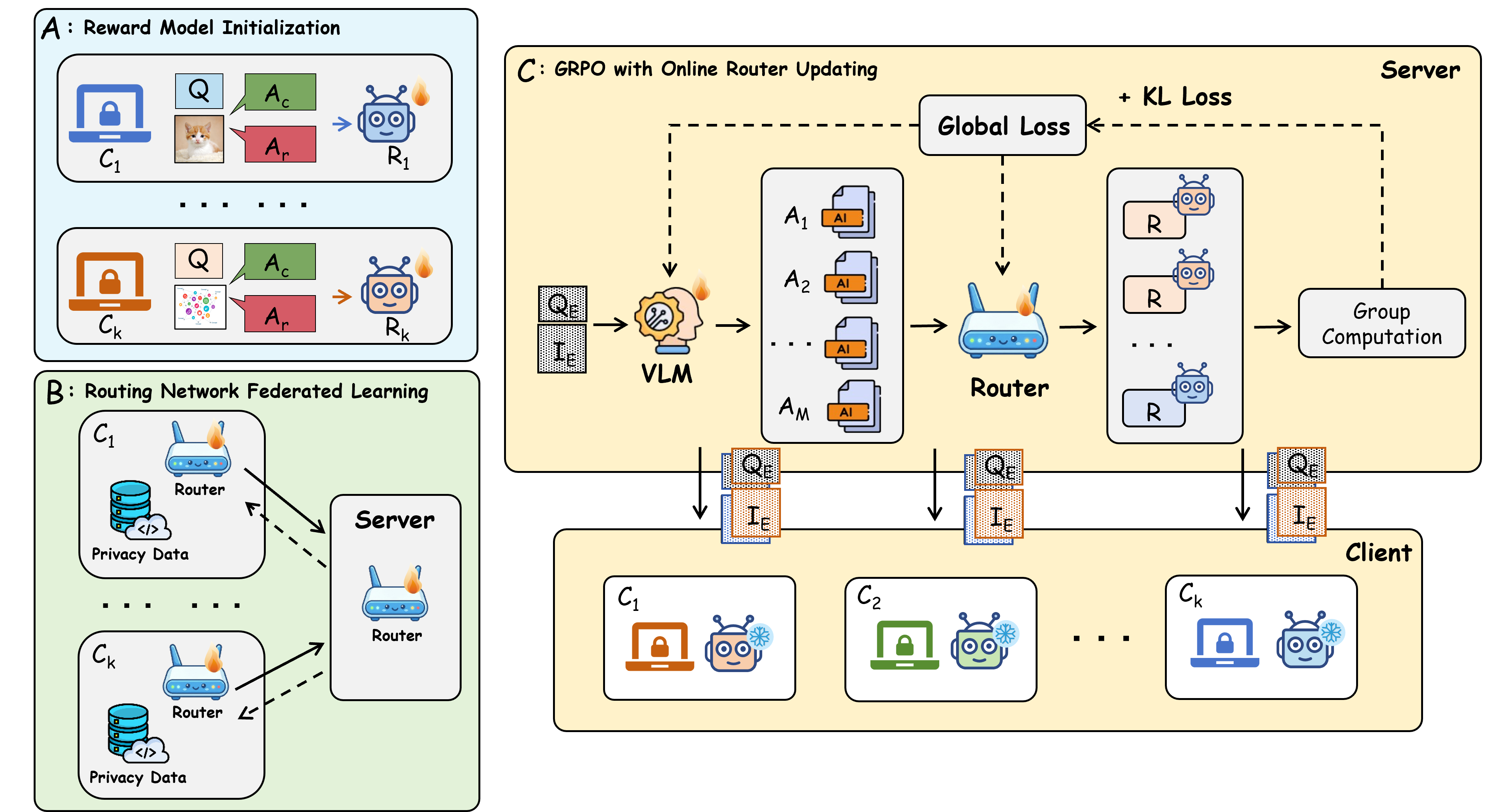} 
    \caption{Overview of our proposed federated framework architecture. The framework is composed of three interconnected stages: (A) Reward Model Initialization: Individual clients independently train local reward models (RMs) on their private, domain-specific datasets. (B) Routing Network Federated Learning: A global routing mechanism is collaboratively trained across clients to learn how to dispatch inputs to the most capable local RM. (C) GRPO Policy Alignment: The final multimodal policy is optimized using Group Relative Policy Optimization (GRPO), guided by the high-quality reward signals selected by the trained router.}
    \label{fig:framework}
\end{figure*}

\section{Related Works}
\subsection{LLM Routing and Model Selection}
Routing and model selection have been widely explored in large language model systems as effective mechanisms for improving efficiency, scalability, and robustness. Early studies primarily focus on inference-time routing, where routers dynamically select experts or models of different capacities to balance computational cost and performance. A variety of mechanisms have been proposed to steer these routing decisions \cite{lu2024routing,ding2024hybrid,ongroutellm}. And more recent work \cite{frick2025prompt} further improves the scalability of inference routing, and extends inference-time routing to multimodal large models \cite{tang2025ecvl}.
Recently, routing mechanisms have been introduced into the reward modeling stage of RLHF. In this line of work, reward model selection is often formulated as a multi-armed bandit problem, where different reward models correspond to distinct arms. For instance, algorithms like LinUCB \cite{li2025llm} and Bayesian Thompson Sampling \cite{wu2025reward} have been employed to select optimal reward models or guide routing based on training signals. However, existing reward routing approaches are typically designed under centralized assumptions and are restricted to text-only models, without considering federated learning scenarios with heterogeneous reward models distributed across different clients, nor extending routing-based reward selection to multimodal alignment settings.

\subsection{Decentralized and Federated Preference Alignment}
Most existing preference alignment methods are developed under centralized assumptions, where human feedback and reward models are collected and optimized within a single training pipeline \cite{zhao2023group,ramesh2024group,chakraborty2024maxmin}. However, such centralized settings are often impractical in scenarios involving privacy-sensitive data or distributed data ownership. To address these concerns, FL is introduced into preference alignment, enabling collaborative model training without sharing raw preference data. Recent work \cite{srewa2025pluralllm} extends GPO frameworks to federated settings, allowing multiple clients or groups to jointly learn preference predictors while preserving data locality. Subsequent study \cite{srewa2025systematic} further introduces dynamic preference selection strategies and extends these frameworks to reinforcement learning-based training. Another approach \cite{wu2024client} mitigates preference heterogeneity by clustering or aggregating client-side preference models. However, existing decentralized preference alignment methods often incur high computational and communication costs and remain highly sensitive to heterogeneous preference distributions across clients. Moreover, these methods typically focus on static preference aggregation or selector learning, without jointly optimizing reinforcement learning policies under dynamically selected or routed reward signals. This disconnection limits their ability to adapt effectively to complex, varying client preferences in scalable real-world applications.

\section{Task Definition}
\label{sec:task definition}
In this work, we address a preference alignment problem under a federated and heterogeneous VLMs setting. Let $\mathcal{C}=\{1,\dots,K\}$ denote the set of clients. Let $\pi_{\theta}(y\mid x)$ denote a VLM policy parameterized by $\theta$, which generates an output $y\in\mathcal{Y}$ given a multimodal input $x\in\mathcal{X}$. Each client $k\in\mathcal{C}$ owns a private preference dataset $\mathcal{D}_k^{\mathrm{pref}}$, consisting of multimodal inputs and pairwise preference annotations:
\[
\mathcal{D}_k^{\mathrm{pref}}=\{(x_i^k,y_i^{k,+},y_i^{k,-})\},
\]
where $y_i^{k,+}\succ y_i^{k,-}$ indicates that $y_i^{k,+}$ is preferred over $y_i^{k,-}$ under the local annotation criterion. The client data are non-IID, i.e., $\mathcal{D}_k^{\mathrm{pref}}\sim\mathcal{P}_k$ and $\mathcal{P}_k\neq \mathcal{P}_{k'}$ for $k\neq k'$, where heterogeneity arises from differences in visual content, linguistic expressions, and local preference criteria across clients.

Specifically, we focus on vision-language tasks and take Visual Question Answering (VQA) as a representative example. In addition to private client preference data, the central server maintains a public dataset. 
Each client is associated with an implicit preference function $R_k:\mathcal{X}\times\mathcal{Y}\rightarrow\mathbb{R}$, which assigns higher scores to outputs that better match its local evaluation criteria. 
The goal of federated preference alignment is to optimize the VLM policy such that it maximizes the expected mixed reward over public query inputs:
\[
\theta^*=\arg\max_{\theta}\,\mathbb{E}_{x\sim\mathcal{D}^{\mathrm{pub,qry}},\,y\sim\pi_{\theta}(\cdot\mid x)}
\bigl[R_{\mathrm{mix}}(x,y)\bigr].
\]

Due to FL privacy constraints, raw client data, preference annotations, and reward model parameters never leave local clients.

\section{Method}
\subsection{Framework Overview}

To address heterogeneous reward coordination in privacy-sensitive federated VLM alignment, our framework (Figure \ref{fig:framework}) employs a three-stage decentralized process that keeps sensitive labels and reward model parameters strictly on-device. First, to capture unique, domain-specific evaluation criteria, each client independently trains a local reward model on its private preference data, ensuring that sensitive labels and reward model parameters remain strictly on-device. 
Then, the server broadcasts a lightweight routing network $g_{\phi}$ to participating clients for federated optimization. Each client updates the router locally using its private preference data and on-device reward model, and the server aggregates the returned router updates to obtain a global routing mechanism.
Finally, during the alignment phase, the server uses the router to identify the most relevant client for specific query-response pairs. The base VLM is optimized using Group Relative Policy Optimization (GRPO) based on the returned scalar rewards. The routing network is updated online to track distribution shifts as the policy model evolves.

\subsection{Reward Model Initialization}
\label{sec:rm}
Each client independently initializes and trains a local reward model based on its private multimodal preference dataset $\mathcal{D}^{\mathrm{pref}}_k$, as shown in Figure \ref{fig:framework}(A). The reward model $R_k$ maps a multimodal input–output pair $(x, y)$ to a scalar score $R_k(x, y)$, reflecting the alignment between the candidate output and the client-specific preferences.
For training, each preference tuple $(x_i^k, y_i^{k,+}, y_i^{k,-}) \in \mathcal{D}^{\mathrm{pref}}_k$ indicates that $y_i^{k,+}$ is preferred over $y_i^{k,-}$. Following standard reward modeling practice, we adopt a pairwise preference learning objective grounded in the Bradley–Terry model \cite{rm_rule}, which can be expressed as:
$$\mathcal{L}_{\mathrm{RM}}^{(k)} = - \mathbb{E}_{\mathcal{D}_k^{pref}} \left[ \log \sigma \!\ \Delta R_k(x_i^k, y_i^{k,+}, y_i^{k,-}) \right],$$
where $\sigma(\cdot)$ denotes the logistic sigmoid function.

To ensure reward signals are comparable across heterogeneous clients, we apply local normalization. Each client computes the empirical mean $\mu_k$ and standard deviation $\sigma_k$ from its local training set. When queried by the server, the client returns a standardized score:
$$\bar {R}_k(x,y) = \frac{R_k(x,y)-\mu_k}{\sigma_k+\epsilon}.$$
This standardization mitigates scale variation and stabilizes the server's routed reward aggregation.

\subsection{Routing Network Federated Learning}
\label{sec:router}

To coordinate heterogeneous client-specific reward models under federated constraints, we train the routing network $g_{\phi}$ in a standard federated manner. After local reward-model training, each client keeps its reward model $R_k$ on-device and uses it together with its private preference data only for local router optimization. The server initializes the global router parameters $\phi^{(0)}$ and broadcasts them to participating clients at each communication round. Each selected client then updates the router locally and sends only router parameter updates back to the server, which aggregates them to form the next global router.

Concretely, for each input--response pair $(x,y)$, we first obtain a multimodal representation
\[
h = \mathrm{Enc}(x,y),
\]
where $\mathrm{Enc}(\cdot)$ denotes a frozen multimodal encoder. The router then maps $h$ to a logit vector over all clients:
\[
z = g_{\phi}(h) \in \mathbb{R}^{K},
\]
where the $j$-th component $z_j$ indicates the routing score assigned to client-specific reward model $R_j$.

Since each local preference tuple is sampled from client $k$, both the chosen response $y^{k,+}$ and the rejected response $y^{k,-}$ reflect the same client-specific preference source. Therefore, during local router training on client $k$, we supervise the router to assign both responses to expert $k$. Let $e_k \in \{0,1\}^{K}$ denote the one-hot target vector whose $k$-th entry equals $1$. For a local preference tuple $(x_i^k, y_{i}^{k,+}, y_{i}^{k,-})$, the local routing loss is defined as
\[
\mathcal{L}_{\text{router}}^{(k)}
=
\frac{1}{2}
\Big[
\ell_{\mathrm{BCE}}(z_i^{k,+}, e_k)
+
\ell_{\mathrm{BCE}}(z_i^{k,-}, e_k)
\Big],
\]
where $z_i^{k,+}$ and $z_i^{k,-}$ denote the router logits for the chosen and rejected responses, respectively, and $\ell_{\mathrm{BCE}}$ is the binary cross-entropy loss with logits.
Intuitively, this objective encourages the router to learn which client-specific reward model is most relevant for samples drawn from each client distribution.

Each participating client performs several local optimization steps on $\mathcal{L}_{\text{router}}^{(k)}$ and returns the updated router parameters $\phi_k$ to the server. The server then aggregates the local router updates using standard federated averaging:
\[
\phi^{(t+1)}
=
\sum_{k \in \mathcal{S}_t}
\frac{n_k}{\sum_{j \in \mathcal{S}_t} n_j}\,\phi_k^{(t+1)},
\]
where $\mathcal{S}_t$ denotes the set of participating clients at round $t$, and $n_k$ is the number of local training samples on client $k$.

\subsection{GRPO with Online Router Updating}
\label{sec:grpo}
\subsubsection{GRPO with MoR}
After router training, we perform server-side preference alignment of the VLM policy using GRPO. Let $\pi_{\theta_t}$ denote the current policy at iteration $t$, with $\pi_{\theta_0}$ initialized from the base VLM. At each iteration, the server samples a mini-batch of public query inputs $\{x_i\}_{i=1}^{B}\subset \mathcal{D}^{\mathrm{pub,qry}}$. For each input $x_i$, the current policy generates a group of $M$ candidate responses
\[
\{y_{i,j}\}_{j=1}^{M}\sim \pi_{\theta_t}(\cdot\mid x_i).
\]

For each input--response pair $(x_i,y_{i,j})$, the server feeds the pair into the routing network. The router produces routing weights over the client-local reward models:
\[
\boldsymbol{\alpha}_{i,j}
=
\mathrm{softmax}\!\left(g_{\phi}(x_i,y_{i,j})\right),
\]
where $\alpha_{i,j}^{(k)}$ denotes the weight assigned to reward model $R_k$. Accordingly, the mixed reward is defined as
\[
R_{\mathrm{mix}}(x_i,y_{i,j};\phi)
=
\sum_{k=1}^{K}\alpha_{i,j}^{(k)}\,\bar{R}_k(x_i,y_{i,j}).
\]

In practice, to reduce online communication and computation, we employ sparse routing during alignment. Specifically, for each pair $(x_i,y_{i,j})$, the router selects the single most suitable client reward model, and the server queries only this selected client for the scalar reward score. This returned score directly instantiates the mixed reward for policy optimization.

Based on the routed rewards, we compute a group-relative advantage $\hat{A}_{i,j}$ for each candidate response. Let $\pi_{\theta_{\mathrm{old}}}$ denote the policy before the current update step, and let $\pi_{\mathrm{ref}}$ denote the KL-regularized reference policy. The GRPO objective is:
\[
\mathcal{J}_{\mathrm{GRPO}}(\theta)
=
\mathbb{E}_{x_i,\,y_{i,j}}
\big[
\mathcal{L}_{\mathrm{clip}}(\theta,\hat{A}_{i,j})
-
\beta\,\mathcal{L}_{\mathrm{KL}}(\theta)
\big],
\]
where $x_i \sim \mathcal{D}^{\mathrm{pub,qry}}$, 
$y_{i,j} \sim \pi_{\theta_{\mathrm{old}}}(\cdot \mid x_i)$, and
$\mathcal{L}_{\mathrm{KL}}(\theta)$ represents KL loss and $\beta$ regulates KL strength. By optimizing this objective, the policy is encouraged to increase the likelihood of responses that receive higher routed rewards while remaining close to the reference policy.

\subsubsection{Online Router Updating}

The above formulation assumes a fixed routing function during policy optimization. In practice, however, the router is trained on static preference data, while the policy $\pi_\theta$ keeps evolving during GRPO, causing a distribution mismatch between offline router training and on-policy responses. To mitigate this mismatch, we update the router online during GRPO training.

Specifically, we cast online reward routing as a contextual bandit problem \cite{li2025llm,wu2025reward}. For each generated input--response pair $(x_t, y_t)$, we construct a context embedding $c_t=\mathrm{Enc}(x_t,y_t)$, and treat each client-specific reward model $R_k$ as one bandit arm. The router $g_\phi$ outputs a score vector, where $\hat r_k(c_t;\phi)$ estimates the utility of selecting reward model $R_k$ for the current context. We then adopt Neural Thompson Sampling \cite{zhang2021neural} to balance exploration and exploitation under the non-stationary response distribution.

To encourage exploration, Neural Thompson Sampling constructs a Gaussian posterior over the predicted reward for each arm:
$$\tilde r_{t,k} \sim \mathcal N\!\big(\hat r_k(c_t;\phi_{t-1}),\;\nu^2 \sigma^2_{t,k}\big),$$
where $\sigma^2_{t,k}$ represents the exploration variance for arm $k$ at time $t$, which can be computed as
$$\quad\sigma^2_{t,k}=\lambda \,\nabla_\phi \hat r_k(c_t;\phi_{t-1})^\top\mathbf U_{t-1}^{-1}\nabla_\phi \hat r_k(c_t;\phi_{t-1}),$$
where $\lambda$ is a regularization coefficient that controls the prior uncertainty and scales the exploration variance in Neural Thompson Sampling, and $\mathbf U_t$ is an empirical covariance matrix that accumulates gradient information over time, and $\nu$ controls the exploration scale.
The routing decision is made by selecting the arm with the highest sampled reward:
    $$k_t = \arg\max_k \tilde r_{t,k}.$$
After selecting reward model $R_{k_t}$, the router observes a scalar bandit feedback derived from the change in the global GRPO objective:
\[\Delta \mathcal{J}_{t} =  \mathcal{J}_{\text{GRPO}}^{t}- \mathcal{J}_{\text{GRPO}}^{t-1},\]
then the feedback signal is defined as a binary reward
$$r_t = 
\begin{cases} 
0, & \text{if } \Delta \mathcal{J}_{t} > 0 \\ 
1, & \text{if } \Delta \mathcal{J}_{t} \leq 0 
\end{cases},$$
indicating whether the selected reward model leads to an improvement in policy optimization.
The routing parameters are then updated by minimizing an $\ell_2$-regularized squared loss:
$$\mathcal L_{\text{online}}(\phi)=\frac{1}{2}\big(\hat r_{k_t}(c_t;\phi) - r_t\big)^2+\frac{\lambda}{2}\|\phi - \phi_0\|_2^2,$$
where $\phi_0$ denotes the router initialization obtained from offline training on static preference data.
Simultaneously, the covariance matrix is updated as
$$
\mathbf U_t=\mathbf U_{t-1}+\nabla_\phi \hat r_{k_t}(c_t;\phi_t)\nabla_\phi \hat r_{k_t}(c_t;\phi_t)^\top.
$$

\section{Experiments}
\subsection{Experimental Settings}
\subsubsection{Dataset.}  We construct a heterogeneous vision--language preference dataset from existing sources including POVID \cite{zhou2024aligning}, VLFeedback \cite{2023vlfeedback}, and MME-RealWorld \cite{zhangmme}. Since these datasets provide supervision in different forms, we first unify them into a pairwise preference format. Specifically, VLFeedback provides score-based annotations over helpfulness, visual faithfulness, and ethical considerations, which we convert into pairwise preferences by comparing candidate responses under the same image--query input. MME-RealWorld is a multiple-choice benchmark, which we transform into preference pairs by treating the correct option as preferred and incorrect options as rejected.
To model realistic data heterogeneity, we construct federated clients at two levels. From POVID and selected VLFeedback subsets, we build three domain-specific clients: Detail Description, Medical Vision-Language Understanding, and OCR-like Visual Reasoning. In addition, we split MME-RealWorld by subtasks and treat each subtask as an individual client. This setup introduces both domain-level and task-level heterogeneity across clients.

\subsubsection{Baselines.} We compare MoR against representative baselines for decentralized preference-based vision--language alignment, including \textbf{Single RM}, which performs alignment independently with only local reward models; \textbf{Random Selection}, which randomly chooses a client reward model for response evaluation; \textbf{Avg RM}, which averages rewards from all client-specific reward models; \textbf{FedAvg} \cite{mcmahan2017communication}, which aggregates reward models by parameter averaging; and \textbf{PluralLLM-alpha} \cite{srewa2025systematic}, which improves the aggregation strategy of PluralLLM \cite{srewa2025pluralllm}, but still assumes a homogeneous reward structure and uses a single global alignment signal without client-aware routing.

\subsubsection{Metrics.} We evaluate open-ended responses using an LLM-as-a-Judge protocol. For each image--instruction input, the judge compares candidate responses along three complementary dimensions: \textit{Helpfulness}, \textit{Visual Faithfulness}, and \textit{Ethical Considerations}. We report pairwise preference win-rates, average judge scores across these three dimensions, and Visual Faithfulness scores. Domain-wise win-rates are reported for the detailed description, medical, and OCR-like subsets. For MME-RealWorld, which is a multiple-choice benchmark, we directly report answer accuracy (\textbf{Acc.}). To improve robustness against stochastic decoding, we adopt a best-of-3 sampling strategy for all evaluated policies under identical decoding settings. The detailed LLM-as-a-Judge prompt template is provided in the Supplementary Material.

\subsubsection{Implementation Details.} 
We use Qwen2.5-VL-7B-Instruct \cite{qwen2.5-VL} as the common initialization model in all experiments to eliminate potential confounding factors.
During the reinforcement learning stage, we adopt a modified EasyRL framework \cite{zheng2025easyr1}.
While the vision tower remains frozen across the entire experimental pipeline, only the language-side policy parameters are updated during the reinforcement process.
Architecturally, the router is built upon four Transformer layers with a linear classification head. In the online update phase, optimization is restricted to the linear head to maintain the pre-trained representational capacity of the Transformer layers. All experiments are conducted on a server equipped with RTX A6000 PRO GPUs, each with 96GB memory. More implementation details are provided in the Appendix.

\subsection{Main Results}

\begin{table*}[t]
    \centering
    \caption{Main results and cross-category generalization on MME-RealWorld under the 3-client homogeneous-RM setting. All values are accuracy (\%). Diagram, Book Map Poster (Book.), and Position are client-specific in-distribution evaluation subsets, while the remaining categories are held-out MME-RealWorld categories used for cross-category generalization evaluation. The best result in each column is bolded.}
    \label{tab:large_results_split}
    \setlength{\tabcolsep}{2.2pt} 
    \footnotesize 
    \begin{tabularx}{\textwidth}{@{} l ccc c | *{8}{X} c @{}}
        \toprule
        \textbf{Method} & Diagram & Book. & Position & ID Avg. & Attr. & Obj. & Adver. & Color & Count & License & Table & Text. & Overall \\
        \midrule
        \rowcolor{gray!15}
        Qwen2.5-VL-7B    & 83.75 & 87.18 & 58.27 & 76.40 & 27.71 & 27.03 & 86.62 & 57.14 & 16.94 & 81.40 & 79.77 & 80.99 & 62.44 \\
        Best Single      & 60.00 & 84.62 & 50.39 & 65.00 & 39.76 & 48.65 & 82.80 & 50.79 & 28.23 & 81.40 & 56.32 & 78.51 & 60.13 \\
        AvgRM            & 62.50 & 77.56 & 48.03 & 62.70 & 37.35 & \textbf{59.46} & 84.71 & 59.52 & 24.19 & 80.23 & 63.45 & 75.21 & 61.11 \\
        Random           & 50.00 & 59.62 & 51.97 & 53.86 & 40.96 & 41.44 & 61.15 & 53.97 & 31.45 & 56.98 & 42.76 & 59.50 & 49.98 \\
        FedAvg           & 80.63 & 89.10 & 39.37 & 69.70 & 27.71 & 44.14 & 85.99 & 61.11 & 25.00 & 81.40 & 73.10 & \textbf{82.64} & 62.74 \\
        PluralLLM-alpha        & 63.12 & 75.64 & 21.26 & 53.34 & 24.10 & 22.52 & 64.97 & 34.92 & 33.06 & 66.28 & 60.46 & 63.64 & 48.18 \\
        \midrule
        MoR             & \textbf{84.38} & \textbf{91.03} & \textbf{64.56} & \textbf{79.99} & \textbf{42.17} & 51.35 & \textbf{89.17} & \textbf{67.46} & \textbf{33.87} & \textbf{84.88} & 79.31 & \textbf{82.64} & \textbf{70.07} \\
        w/o Online       & 82.50 & 83.97 & 38.58 & 68.35 & 33.73 & 36.04 & 85.35 & 62.70 & 26.61 & \textbf{84.88} & \textbf{79.77} & 76.86 & 62.82 \\
        w/o Norm.         & 66.25 & 78.21 & 41.73 & 62.06 & 28.92 & 24.32 & 82.17 & 26.98 & 25.81 & 76.74 & 66.44 & 78.51 & 54.19 \\
        \bottomrule
    \end{tabularx}
\end{table*}

\subsubsection{Comparison with Federated Reward Baselines.}
We first evaluate MoR under a controlled 3-client homogeneous-RM setting constructed from MME-RealWorld subtasks. 
As shown in Table~\ref{tab:large_results_split}, MoR achieves the best in-distribution average accuracy among all compared methods, reaching 79.99\% on ID Avg. 
Compared with FedAvg, MoR improves the average accuracy by 10.29 percentage points. 
It also outperforms Best Single, AvgRM, Random selection, and PluralLLM by 14.99, 17.29, 26.13, and 26.65 percentage points.

The improvements are consistent across the three client-specific evaluation subsets. 
MoR obtains 84.38\%, 91.03\%, and 64.56\% accuracy on Diagram, Book., and Position, respectively, achieving the best result on all three subsets. 
In contrast, static reward aggregation methods exhibit unstable behavior across clients. 
For example, FedAvg performs competitively on Book. but drops substantially on Position, suggesting that directly averaging reward models can obscure client-specific reward expertise when preference distributions differ. 
AvgRM and Random selection further underperform MoR, indicating that neither uniformly mixing rewards nor selecting rewards without input-dependent routing can effectively resolve federated reward heterogeneity.

These results demonstrate that MoR provides a more effective mechanism for coordinating decentralized reward models than static aggregation or selection-based baselines. 
Instead of forcing heterogeneous reward signals into a single global reward model, MoR adaptively routes each input-response pair to the most suitable client reward model, thereby preserving client specialization while enabling collaborative policy optimization.

\subsubsection{Cross-Category Generalization.}

\begin{table}[!t]
    \centering
    \caption{Performance under heterogeneous reward-model configurations on MME-RealWorld. 
    The three clients use different VLM backbones: Qwen2-VL-2B-Instruct \cite{Qwen2VL}, Qwen2.5-VL-3B-Instruct, and LLaVA-OneVision-0.5B-Instruct \cite{li2024llavaonevisioneasyvisualtask}. 
    Best Single denotes the single reward model trained on Client2. 
    All values are accuracy (\%).}
    \label{tab:model_heterogeneity}
    \setlength{\tabcolsep}{4pt} 
    \begin{tabular}{lcccc}
        \toprule
        \setlength{\tabcolsep}{3pt}
        \textbf{Method} & {\small \textbf{Client1}} & {\small \textbf{Client2}} & {\small \textbf{Client3}} & {\small \textbf{Avg.}} \\
        \midrule
        Best Single & 76.25 & \textbf{89.10} & 62.99 & 76.11  \\
        AvgRM & 76.88 & 82.69 & 62.99 & 74.19  \\
        Random & 76.88 & 83.33 & 42.52 & 67.58  \\
        MoR & \textbf{81.87} & 87.18 & \textbf{64.57} & \textbf{77.87} \\
        \bottomrule
    \end{tabular}
\end{table}

Beyond the client-specific evaluation subsets, we examine whether the learned router can generalize to unseen MME-RealWorld categories. 
Specifically, we evaluate aligned policies on eight held-out categories with more than 800 samples, including Attribute motion multiVehicles (Attr.), Object identify (Obj.), Adver and product (Adver.), Color, Count, License, Table, and Text recognition (Text.).

As shown in Table~\ref{tab:large_results_split}, MoR achieves the best overall accuracy of 70.07\%, outperforming the base Qwen2.5-VL-7B model and all federated reward baselines. 
Averaged over the held-out categories, MoR obtains 66.36\% accuracy, exceeding AvgRM and FedAvg by 5.84 and 6.22 percentage points, respectively. 
These results indicate that MoR does not simply overfit to client identities or in-distribution subsets. 
Instead, the router learns transferable associations between multimodal input-response contexts and client reward expertise, enabling effective reward selection even for unseen categories.

\begin{figure}[t]
    \centering
    \includegraphics[width=\linewidth]{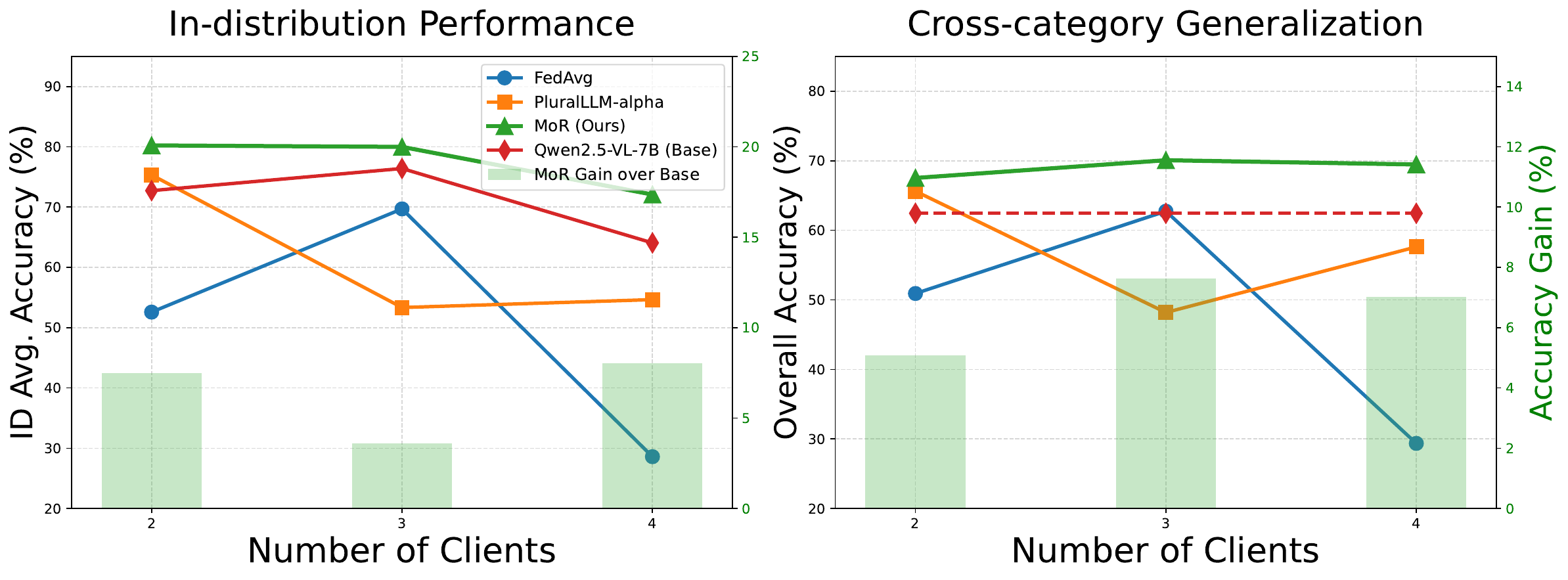}
    \caption{
    Scaling performance with more clients on MME-RealWorld. 
    Left: in-distribution average accuracy over client-specific evaluation subsets. 
    Right: overall accuracy on MME-RealWorld categories. 
    The green bars indicate the accuracy gain of MoR over the base Qwen2.5-VL-7B model. 
    }
    \label{fig:scaling_clients}
\end{figure}

\subsubsection{Scaling to More Clients.}
We further evaluate the scalability of MoR by increasing the number of participating clients from 2 to 4, with all clients constructed from MME-RealWorld subtasks and evaluated using the same metrics. 
As shown in Figure~\ref{fig:scaling_clients}, MoR maintains strong performance as the number of clients grows. 
On in-distribution evaluation, MoR achieves the best ID average accuracy in the 2-client and 3-client settings, and still clearly outperforms FedAvg and PluralLLM-alpha under the 4-client setting. 
On cross-category generalization, MoR consistently obtains the highest overall accuracy across all client numbers, while aggregation-based baselines show larger fluctuations. 
These results indicate that MoR can scale to more heterogeneous clients by adaptively routing each input-response pair to the most suitable client reward model, rather than relying on static reward aggregation.

\subsection{Robustness to Heterogeneity}
\subsubsection{Model heterogeneity.}
As shown in Table~\ref{tab:model_heterogeneity}, MoR achieves the best average accuracy of 77.87\%. 
Although Best Single, which is trained on Client2, obtains the highest accuracy on Client2, its performance is less stable on the other clients. 
In contrast, MoR achieves the best results on Client1 and Client3 and improves the average accuracy over Best Single, AvgRM, and Random. 
These results indicate that MoR can better coordinate reward models with heterogeneous architectures, rather than relying on a single client-specific model or static reward aggregation.

\subsubsection{Stronger non-IID.}
\begin{figure}[t]
    \centering
    \includegraphics[width=\linewidth]{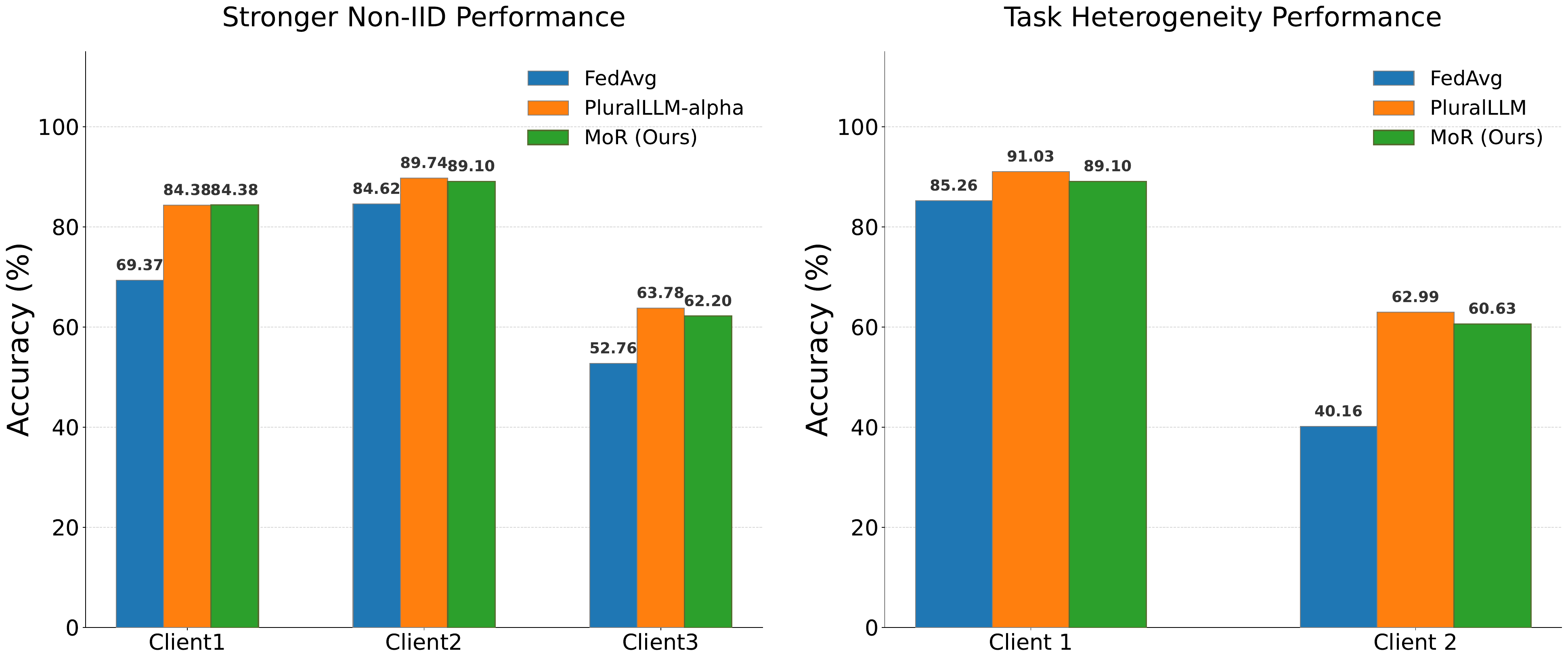}
    \caption{
Performance under stronger distribution and task heterogeneity on MME-RealWorld. 
Left: stronger non-IID setting, where the reinforcement training data are drawn from a distribution different from all client datasets. 
Right: task heterogeneity setting, where MME-RealWorld clients are evaluated under heterogeneous task construction.}
    \label{fig:stronger_noniid}
\end{figure}

Beyond model-level heterogeneity, we also consider a stronger non-IID setting where the reinforcement training data are drawn from a distribution different from all client datasets. 
As shown in Figure~\ref{fig:stronger_noniid}, MoR remains robust under this distribution shift, achieving performance close to PluralLLM and clearly outperforming FedAvg across all three clients. 
PluralLLM obtains slightly higher accuracy, which may be attributed to its aggregation strategy that accesses all client reward models during reward estimation. 
In contrast, MoR adopts sparse routing and queries only the selected client reward model for each input-response pair. 
The comparable performance suggests that MoR can maintain strong robustness to out-of-client training distributions while preserving a more communication-efficient reward coordination mechanism.

\subsubsection{Task heterogeneity.}
\begin{table}[!thbp]
    \centering
    \caption{
    Performance comparison on open-generation vision-language tasks. 
    LLaVA-Med and LLaVAR are used as open-generation clients and evaluated by an LLM-as-a-Judge protocol. 
    Avg. denotes the average judge score, WR. denotes the win rate against the rejected answer, and VF. denotes visual faithfulness. 
    The detailed judge prompt is provided in the Appendix. 
    }
    \label{tab:comparison}
    \setlength{\tabcolsep}{4pt} 
    \begin{tabular}{lcccccc}
        \toprule
        {\textbf{Method}} & \multicolumn{3}{c}{\textbf{llavamed}} & \multicolumn{3}{c}{\textbf{llavar}} \\
        \cmidrule(lr){2-4} \cmidrule(lr){5-7}
        & Avg. & WR. & VF. & Avg. & WR. & VF. \\
        \midrule
        Fedavg    & 6.21 & 85.52 & 5.70 & 3.79 & 36.15 & 2.81 \\
        Plural11m-alpha & 9.12 & 97.96 & 9.03 & 9.52 & 96.27 & 9.45 \\
        MoR      & \textbf{9.36} & \textbf{98.00} & \textbf{9.25} & \textbf{9.81} & \textbf{99.00} & \textbf{9.79} \\
        \bottomrule
    \end{tabular}
\end{table}

We also evaluate MoR under task-level heterogeneity, where clients are constructed from both MME-RealWorld multiple-choice tasks and open-generation vision-language tasks. 
For MME-RealWorld clients, we report accuracy, while for open-generation clients, we use an LLM-as-a-Judge protocol with average score, win rate against the rejected answer, and visual faithfulness.
As shown in Figure~\ref{fig:stronger_noniid} and Table~\ref{tab:comparison}, MoR consistently outperforms FedAvg across both evaluation formats. 
On the open-generation clients, MoR achieves the best results on both LLaVA-Med and LLaVAR across all metrics, indicating stronger response quality and visual grounding. 
On the MME-RealWorld clients, MoR remains competitive with PluralLLM, while both methods substantially outperform FedAvg. 
The slightly higher accuracy of PluralLLM on MME-RealWorld may come from its access to all client reward models during aggregation, whereas MoR relies on sparse routing to query only the selected client reward model. 
These results suggest that MoR can handle heterogeneous task formats while maintaining strong performance with a more selective reward coordination mechanism.

\subsection{Additional Analysis}
\begin{table}[!tbp]
    \centering
    \caption{Router analysis based on pairwise preference ranking accuracy.  
    Router denotes the combination of the learned router and the selected client reward model, while Book., Diagram, and Position denote individual client reward models. 
    ID is the in-distribution evaluation set; Attr. is a non-IID category; Color and Table are same-class but non-overlapping subsets.}
    \label{tab:dataset_comparison}
    \setlength{\tabcolsep}{4pt} 
    \begin{tabular}{lcccc}
        \toprule
        \textbf{dataset} & ID & Attr. & color & table \\
        \midrule
        Router          & \textbf{69.81} & \textbf{62.60} & \textbf{70.23} & \textbf{65.44} \\
        Book.           & 57.67 & 56.51 & 62.99 & 53.42 \\
        Diagram               & 50.11 & 38.01 & 47.26 & 40.90 \\
        Position              & 38.15 & 30.14 & 34.09 & 37.25 \\
        \bottomrule
    \end{tabular}
\end{table}

\subsubsection{Analysis of Router.}
Table~\ref{tab:dataset_comparison} shows the accuracy of different reward-evaluation strategies in judging whether the chosen response is preferred over the rejected response. 
The Router row denotes the combination of the learned router and client reward models, while Book., Diagram, and Position denote using a single client reward model for evaluation. 
Compared with any single reward model, Router achieves the highest accuracy across all evaluation settings, including the in-distribution set, the non-IID set, and the same-class but non-overlapping subsets. 
This indicates that the learned router can select more suitable client-specific reward models for different input-response pairs, leading to more reliable preference judgment than relying on a fixed reward model.

\subsubsection{Analysis of Online Update Effectiveness.}
We evaluate the effectiveness of online router updating by comparing MoR with the variant w/o Online in Table~\ref{tab:large_results_split}. 
Removing online updating leads to a clear performance drop, reducing ID Avg. from 79.99\% to 68.35\% and Overall accuracy from 70.07\% to 62.82\%. 
The degradation is especially pronounced on the Position subset, where accuracy decreases from 64.56\% to 38.58\%. 
This confirms that a router trained only on static preference data can become mismatched with the evolving on-policy response distribution during GRPO. 
By updating the router online, MoR can better adapt reward selection to the changing policy behavior, leading to more stable in-distribution performance and stronger cross-category generalization.

\subsubsection{Analysis of Reward Normalization.}
We further analyze the role of reward normalization by comparing MoR with w/o Norm in Table~\ref{tab:large_results_split}. 
Without normalization, performance drops substantially across both client-specific and held-out categories, with ID Avg. decreasing from 79.99\% to 62.06\% and Overall accuracy decreasing from 70.07\% to 54.19\%. 
This suggests that raw reward scores from different client reward models are not directly comparable, even when the reward models share the same architecture. 
The large degradation on categories such as Color further indicates that scale mismatch can distort routing and policy optimization when reward signals are aggregated or selected across clients. 
Local reward normalization therefore plays an important role in calibrating heterogeneous reward signals and stabilizing MoR under federated reward coordination.

\begin{table}[!t]
    \centering
    \caption{Query-only black-box membership inference results under a threshold attack on reward margins (member:non-member = 1:1). TPR is reported at false positive rates (FPR) of 1\%, 5\%, and 10\%. For all metrics, higher values indicate stronger attack performance (i.e., greater privacy leakage); for AUC, values close to 50\% indicate near-random guessing.}
    \label{tab:mia_results}
    \setlength{\tabcolsep}{1pt} 
    \begin{tabular}{lcccc}
        \toprule
        \setlength{\tabcolsep}{3pt}
        \textbf{Method} & {\small \textbf{AUC (\%)}} & {\small \textbf{TPR@1\%}} & {\small \textbf{TPR@5\%}} & {\small \textbf{TPR@10\%}} \\
        \midrule
        Norm + Clip. 1.5 & \textbf{48.716} & \textbf{0.002} & \textbf{0.042} & \textbf{0.078} \\
        Norm + Clip. 2.0 & 49.778 & 0.003 & 0.053 & 0.090 \\
        Norm                & 49.718 & 0.010 & 0.221 & 0.084 \\
        w/o Norm            & 51.378 & 0.008 & 0.060 & 0.107 \\
        \bottomrule
    \end{tabular}
\end{table}

\subsection{Privacy Attack Evaluation}
\begin{table}[t]
    \centering
    \caption{
    Gradient inversion attack results on transmitted router updates. 
    We report reconstruction quality under different optimization steps. 
    The consistently low PSNR/SSIM and high LPIPS indicate that the attacker fails to recover visually faithful images from router updates.
    }
    \label{tab:gradient_inversion}
    \setlength{\tabcolsep}{6pt}
    \begin{tabular}{lccc}
        \toprule
        \textbf{Steps} & \textbf{PSNR $\uparrow$} & \textbf{SSIM $\uparrow$} & \textbf{LPIPS $\downarrow$} \\
        \midrule
        1000 & 7.671  & 0.0142 & 1.1650 \\
        5000 & 10.665 & 0.0311 & 1.0316 \\
        8000 & 10.096 & 0.0147 & 1.1988 \\
        \bottomrule
    \end{tabular}
\end{table}

We consider a malicious-server threat model and evaluate the privacy exposure of MoR under the information visible to the server. 
Compared with other methods, MoR reduces privacy risks by narrowing the exposed attack surface.
Client preference data and reward-model parameters remain local. 
During router training, the server only receives lightweight router updates, and during GRPO alignment, it only obtains scalar reward scores from selected clients.

We evaluate two attacks. 
First, we conduct a query-only membership inference attack using reward margins between preferred and rejected responses as the membership signal. 
As shown in Table~\ref{tab:mia_results}, all variants achieve AUC values close to 50\%, indicating near-random attack performance and limited membership leakage from scalar reward scores.

Second, since router training involves parameter updates, we evaluate a gradient inversion attack on transmitted router updates. 
As shown in Table~\ref{tab:gradient_inversion}, the reconstructed images have low PSNR and SSIM, together with high LPIPS values across different optimization steps. 
Even after 5000 steps, the attack only achieves 10.665 PSNR and 0.0311 SSIM, suggesting poor reconstruction quality.

\subsection{Communication and Efficiency Analysis}

\begin{figure}[!t] 
\centering
\includegraphics[width=0.45\textwidth]{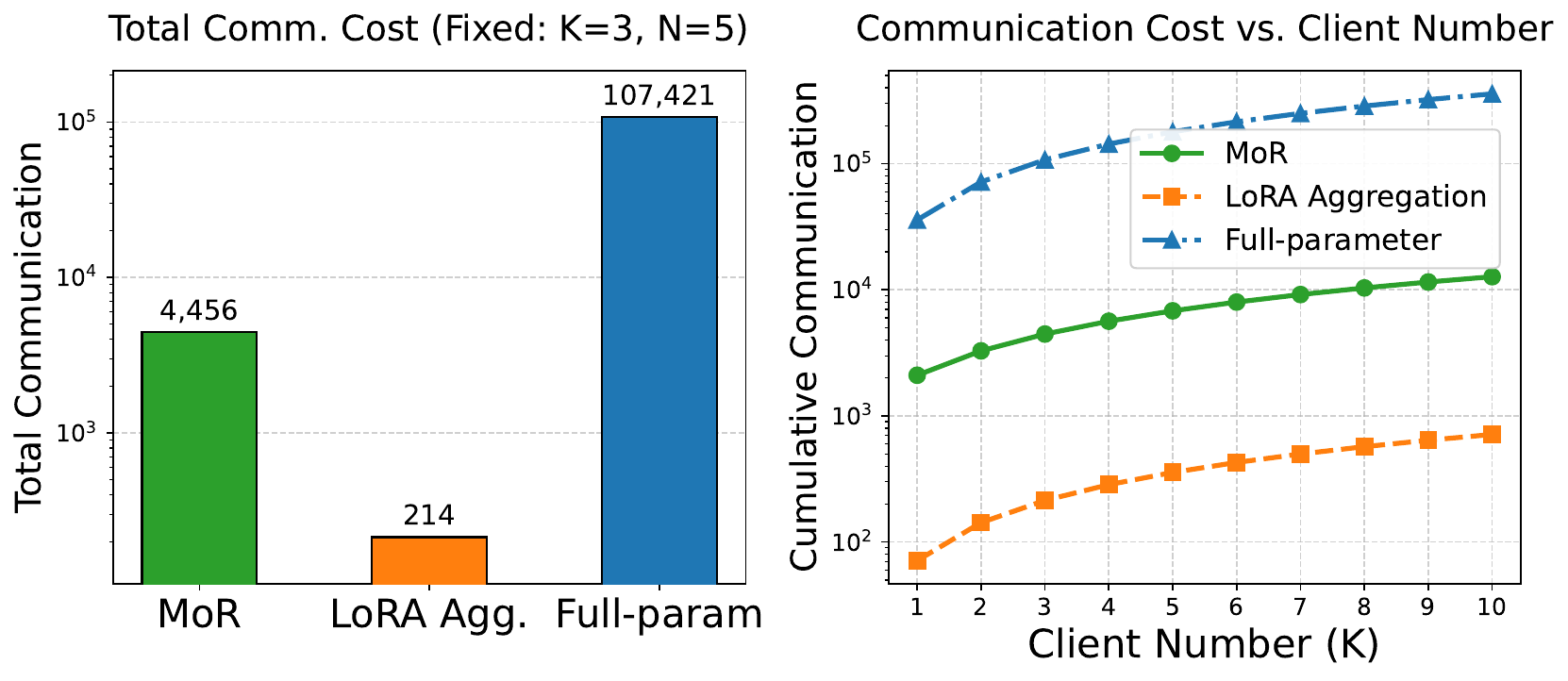} 
\caption{Communication overhead comparison across different aggregation methods. Left figure: Total communication cost (in MB) under a fixed federated setting with $K=3$ clients and $N=5$ training epochs. Right figure: Cumulative communication cost scaling with respect to the number of participating clients ($K$). Note the logarithmic scale on the y-axes. }
\label{fig:communication_cost}
\end{figure}

We analyze the efficiency of MoR from both the communication and computation perspectives. In Stage 1, each client trains its local reward model independently on private preference data, and thus no server--client communication is required. The main communication cost of MoR arises in Stages 2 and 3.

In Stage 2 (router training), MoR follows a standard federated learning protocol for optimizing the routing network. 
Therefore, the communication cost of router training depends on the size of the router rather than the size of client reward models or raw client data. 
Concretely, if $K$ denotes the number of participating clients and $P_\phi$ denotes the number of router parameters, the per-round communication cost is approximately {\boldmath $O(KP_\phi)$}. 
Since the router is lightweight compared with multimodal reward models, this stage introduces much lower communication overhead than synchronizing full reward-model parameters.

In Stage 3 (GRPO-based alignment), the server generates candidate responses on public query inputs and uses the trained router to select the most suitable client reward model for each input--response pair. 
With sparse routing, only one client reward model is queried for each sample. 
If $B$ denotes the number of queried samples per round, $S_q$ the size of each transmitted query pair, and $S_r$ the size of each returned scalar reward, the per-round communication cost becomes {\boldmath $O(B(S_q+S_r))$}. 
Thus, the online alignment communication depends on the number of routed reward queries and returned scalar scores, rather than on the size of reward-model parameters.

By contrast, parameter-aggregation baselines such as FedAvg require repeatedly synchronizing full reward-model parameters between the server and clients, leading to a per-round communication cost of {\boldmath $O(KP)$}, where $P$ denotes the number of synchronized reward-model parameters. 
Since modern multimodal reward models are typically large, such parameter-level synchronization introduces substantial communication overhead. 
As illustrated in Figure~\ref{fig:communication_cost}, MoR is substantially more communication-efficient than full-parameter synchronization because it only federates a lightweight router in Stage 2 and performs sparse reward queries in Stage 3.

At the same time, Figure~\ref{fig:communication_cost} shows that MoR does not necessarily achieve lower raw communication volume than LoRA-based aggregation in the current multimodal setting. 
This is because LoRA transmits compact low-rank adapter updates, whereas MoR still needs to exchange multimodal query payloads during online reward evaluation. 
In our setting, image transmission accounts for a large portion of MoR's communication cost, indicating that the dominant overhead comes from visual payloads rather than from the reward-query protocol or router training. 
This also suggests that MoR's communication cost can be further reduced by practical visual payload optimization, such as feature caching, image compression, or transmitting compact image representations.

From the computation perspective, the additional cost of MoR beyond standard GRPO comes from router inference and client-side reward evaluation. With sparse routing, only a small subset of client reward models is queried for each input--response pair, so the extra inference cost scales with the number of selected experts rather than the total number of clients. In contrast, averaging-based baselines like PluralLLM-alpha \cite{srewa2025systematic} require evaluating all client reward models for each candidate response, leading to higher inference overhead as the number of clients grows.

\section{Conclusion}
In this paper, we propose MoR, a federated alignment framework for heterogeneous vision-language models. Instead of aggregating model parameters, MoR coordinates client-specific preference signals through a routing-based Mixture-of-Rewards mechanism and optimizes the policy with GRPO. This design preserves local reward expertise while avoiding direct sharing of client data or reward-model parameters.
Experiments on diverse vision-language benchmarks show that MoR consistently improves generalization, robustness, and cross-client adaptability over existing federated alignment baselines. Additional analyses further demonstrate its advantages in privacy preservation and communication efficiency. These results suggest that preference-level collaboration is a promising direction for scalable and privacy-preserving alignment of heterogeneous VLMs.

\section{Acknowledgments}
This work was funded by the National Natural Science Foundation of China (NSFC) under Grants No.U25B2070 and No. 62406013, the Beijing Advanced Innovation Center Funds for Future Blockchain and Privacy Computing(GJJ-24-034), and the Fundamental Research Funds for the Central Universities.

\bibliography{aaai2026}

\end{document}